 \documentclass[pmlr,twocolumn,10pt]{jmlr} % W&CP article

\usepackage{longtable}
\usepackage{booktabs}

\usepackage{siunitx}

\usepackage[switch]{lineno}
\usepackage{dirtytalk}
\usepackage{authblk}
\usepackage{algpseudocode}% http://ctan.org/pkg/algorithmicx

\theorembodyfont{\upshape}
\theoremheaderfont{\scshape}
\theorempostheader{:}
\theoremsep{\newline}

\jmlrvolume{LEAVE UNSET}
\jmlryear{2024}
\jmlrsubmitted{LEAVE UNSET}
\jmlrpublished{LEAVE UNSET}
\jmlrworkshop{Machine Learning for Health (ML4H) 2024} % W&CP title

 \title[MedG--KRP: Medical Graph Knowledge Representation Probing]{MedG--KRP: Medical Graph Knowledge Representation Probing}

\author[1]{Gabriel R. Rosenbaum\thanks{Direct all correspondance to \href{mailto:garosenbaum07@gmail.com}{garosenbaum07@gmail.com}}}
\author[3,\textdagger]{Lavender Yao Jiang}
\author[4,\textdagger]{Ivaxi Sheth}
\author[1]{Jaden Stryker}
\author[1,5]{Anton Alyakin}
\author[1,2]{Daniel Alexander Alber}
\author[1,6]{Nicolas K. Goff}
\author[7]{Young Joon (Fred) Kwon}
\author[1,8]{John Markert}
\author[1,9]{Mustafa Nasir--Moin}
\author[1,10]{Jan Moritz Niehues}
\author[1,2]{Karl L. Sangwon}
\author[1,11]{Eunice Yang}
\author[1,3,7,12]{Eric Karl Oermann}

\affil[1]{Department of Neurosurgery, NYU Langone Health, New York, NY, USA}
\affil[2]{New York University Grossman School of Medicine, New York, NY USA}
\affil[3]{Center for Data Science, New York University, New York, NY, USA}
\affil[4]{CISPA Helmholtz Center for Information Security, Saarbrücken, Germany}
\affil[5]{Department of Neurosurgery, Washington University in Saint Louis, Saint Louis, Missouri, USA}
\affil[6]{Department of Neurosurgery, The University of Texas at Austin Dell Medical School, Austin, TX, USA}
\affil[7]{Department of Radiology, NYU Langone Health, New York, NY, USA}
\affil[8]{Heersink School of Medicine, University of Alabama --  Birmingham, Birmingham, AL, USA}
\affil[9]{Harvard Medical School, Boston, MA, USA}
\affil[10]{Else Kroener Fresenius Center for Digital Health, Technical University Dresden, Dresden, Germany}
\affil[11]{Columbia Vagelos College of Physicians \& Surgeons, New York, NY, USA}
\affil[12]{Neuroscience Institute, NYU Langone Heatlh, New York, NY, USA\newline}

\affil[$\dag$]{\emph{Authors contributed equally.}}

\begin{document}

\maketitle

\begin{abstract}

Large language models (LLMs) have recently emerged as powerful tools, finding many medical applications. LLMs' ability to coalesce vast amounts of information from many sources to generate a response-a process similar to that of a human expert---has led many to see potential in deploying LLMs for clinical use. However, medicine is a setting where accurate reasoning is paramount. Many researchers are questioning the effectiveness of multiple choice question answering (MCQA) benchmarks, frequently used to test LLMs. Researchers and clinicians alike must have complete confidence in LLMs' abilities for them to be deployed in a medical setting. To address this need for understanding, we introduce a knowledge graph (KG)--based method to evaluate the biomedical reasoning abilities of LLMs. Essentially, we map how LLMs link medical concepts in order to better understand how they reason. We test GPT-4, Llama3--70b, and PalmyraMed--70b, a specialized medical model. We enlist a panel of medical students to review a total of 60 LLM-generated graphs and compare these graphs to BIOS, a large biomedical KG. We observe GPT-4 to perform best in our human review but worst in our ground truth comparison; vice--versa with PalmyraMed, the medical model. Our work provides a means of visualizing the medical reasoning pathways of LLMs so they can be implemented in clinical settings safely and effectively. 

\end{abstract}

\paragraph*{Data and Code Availability}
Prompts, generated graphs, code and human evaluations are available at \url{https://github.com/nyuolab/MedG-KRP}.

\paragraph*{Institutional Review Board (IRB)}
Our research does not require IRB approval.

\section{Introduction}
\label{sec:intro}
\paragraph{} The increasing use of large language models (LLMs) has diversified their applications beyond standard natural language processing (NLP) tasks such as text generation, translation, and summarization~\citep{Wu2021-eb, OpenAI2023-hq,Dubey2024-qg,Xu2023-in}. The advancements in LLMs' capabilities have led to a growing interest among researchers and healthcare professionals in leveraging LLMs for medical applications ~\citep{Clusmann2023}.  The capacity of LLMs to handle extensive volumes of clinical data, medical records, and scientific literature~\citep{Huang2019-xv, Alsentzer2019-rp, Bolton2022-gy} introduces the potential for advancements in clinical decision support, diagnostics, and patient management~\citep{Yang2022-ki,Jiang2023-ig,Singhal2023-oy, McDuff2023-wk, Tu2024-he}. For safety-critical applications such as healthcare, the performance of LLMs must be vigorously validated ~\citep{Clusmann2023}.

Benchmarking LLMs' medical abilities is a challenging task, however. Medical knowledge, even when limited to common diseases, is vast, making it difficult to design benchmarks that capture the breadth of information clinicians rely on daily~\citep{Jain2024-up}. Additionally, due to the large volume of training data that LLMs memorize, there is concern that their performance on traditional benchmarks may be artificially inflated by memorization~\citep{carlini2023quantifyingmemorizationneurallanguage}. As a result, developing more rigorous and comprehensive benchmarks is essential to accurately evaluate LLMs' true medical understanding and ensure their safe and effective deployment in clinical settings.

The medical capabilities of LLMs are often evaluated through multiple-choice question answering (MCQA) benchmarks~\citep{Pal2022-ui}. Datasets such as MedQA \citep {jin2020USMLEMedQA}, based on questions from the USMLE, draw directly from standardized medical examinations, while other benchmarks like MultiMedQA \citep{Singhal2023LLMsEncodeClinicalKnowledge} aggregate data from a variety of medical knowledge sources.  However, recent findings by \citet{griot2024llmsandmultiplechoicequestions} raise concerns that these MCQA benchmarks may not adequately evaluate the depth of LLMs' medical understanding or reasoning ability, suggesting that performance may be influenced by surface-level pattern recognition rather than genuine clinical reasoning. Moreover, prior studies have demonstrated that some state-of-the-art LLMs exhibit biases in medical reasoning and perform poorly in essential tasks such as medical coding, highlighting further limitations in their practical utility and accuracy \citep{omiye2023llmsrace-basedmedicine, soroush2024llmsarepoormedicalcoders}.

Confidence in LLMs' medical capabilities and the methods used for their evaluation must be ensured before their deployment in clinical settings. It is therefore essential to develop alternative methods for a comprehensive assessment of LLMs' performance.

% Prior work has shown that some state-of-the-art LLMs perpetuate biases in medical reasoning and show poor performance in medical coding, a basic administrative task \cite{omiye2023llmsrace-basedmedicine} \cite{soroush2024llmsarepoormedicalcoders}. Most worryingly, recent work by \citet{griot2024llmsandmultiplechoicequestions} has suggested that the current MCQA benchmarks may simply not reflect LLMs' medical understanding or reasoning ability.

% Natural language processing (NLP) researchers and clinicians alike must have full confidence in LLMs' medical abilities, and thereby the methods used to benchmark these abilities, before LLMs are applied in real world clinical settings. If the effectiveness of MCQA benchmarks have come under question, researchers must create alternative methods in order to understand LLMs' behavior as thoroughly as possible.

Our research is guided by the objective of increasing the transparency of LLMs by structuring their medical reasoning processes. This approach aims to offer a deeper understanding of LLMs' medical performance that extends beyond the capabilities of traditional MCQA benchmarks. To address the limitations inherent in existing evaluation methods, we propose a novel technique for visualizing the connections between medical concepts and understanding pathways in medicine by generating knowledge graphs (KGs). This method reduces the risk of LLMs relying on verbatim memorization from pretraining data and circumvents issues related to the overlap of benchmark data with the training corpus.

% Biomedical knowledge graphs are designed to integrate and categorize extensive medical concepts and their interrelationships. One of the most popular, biomedical KG, is the Unified Medical Language System (UMLS), which categorizes hundreds of thousands of medical concepts and millions of relationships between these concepts \citep{bodenreider2004umls}. However, \citet{Chandak2023primekg} highlighted that it includes 192 subtypes of autism, despite only three being recognized in current clinical practice. We found issues such as incorrectly spelled nodes, erroneous or misleading edges, and unused relationship types. 
% Therefore, PrimeKG, \cite{Chandak2023primekg}, has found success by combining a variety of sources. Others, such as the Biomedical Informatics Ontology System (BIOS), \cite{yu2022biosalgorithmicallygeneratedbiomedicalkg}, utilize ML techniques in generation. Some researchers have also generated KGs by extracting concepts directly from electronic medical records using LMs \cite{Rotmensch2017learningKGs}.

Our work is motivated by the seemingly contradictory nature of LLMs' application. They exhibit the potential to automate complex medical tasks but also present challenges due to their black-box nature and susceptibility to errors. Our approach \emph{MedG--KRP} leverages LLMs to systematically structure and visualize their parametric knowledge. We begin with a single medical concept and use the LLM to identify and generate a knowledge graph of ``causes" and ``effects" associated with this concept. To the best of our knowledge, this is the first work to leverage an LLM to systematically generate a knowledge graph from a single, specified medical concept.

The knowledge graphs generated by the \emph{MedG--KRP} process offer several potential applications. By interpreting these graphs as proxies for LLMs' internal knowledge structures, we can enhance the interpretability of the models by examining their grasp of medical pathways. Additionally, LLM-generated graphs could be employed to augment or correct existing biomedical knowledge graphs. Future research could also explore using \emph{MedG--KRP} for chain-of-thought (COT) prompting, as proposed by \citet{wei2023chainofthoughtpromptingelicitsreasoning}. In this approach, models could first generate knowledge graphs to inform their reasoning process when addressing medical questions.

We generate a total of sixty graphs for twenty medical concepts using three LLMs: GPT-4, Llama3-70b, and PalmyraMed. We enlist a panel of medical experts to score each graph in terms of accuracy and comprehensiveness to the current medical literature. Additionally, we benchmark our graphs against the BIOS KG~\citep{yu2022biosalgorithmicallygeneratedbiomedicalkg} as ground truth. The results from expert evaluation indicate that the accuracy of the generated graphs is generally higher than their comprehensiveness. Additionally, both generalist and specialized medical models show a tendency to incorporate public knowledge, which may influence the graphs' content and affect the representation of clinical information.
% We find that model outputs are often scored by reviewers as between \say{mostly accurate} and \say{completely accurate,} and \say{mostly comprehensive} and \say{completely comprehensive.} We find that average accuracy scores are consistently higher than average comprehensiveness scores. We find an observable influence of public (as opposed to clinical) knowledge in the graphs generated by both generalist and medical models. 

Our contributions can be summarized as follows:
\begin{itemize}
    \item We propose \emph{MedG--KRP} to map the medical knowledge embedded in LLMs, aiming to enhance their explainability.
    \item We analyze LLMs' understanding of causal pathways in medicine by utilizing both human reviewers and comparison to a current biomedical KG.
    \item We observe that medical fine-tuned models performed unexpectedly worse than standard generalist models in human evaluations, despite being specialized for the domain.
    \item We propose possible ways in which our method could be used to repair incomplete KGs and augment traditional COT prompting.

\end{itemize}

\section{Related Works}
\paragraph{Biomedical knowledge graphs} are designed to integrate and categorize extensive medical concepts and their interrelationships. \citet{bodenreider2004umls} proposed the Unified Medical Language System (UMLS), which categorizes hundreds of thousands of medical concepts and millions of relationships between these concepts. Biomedical KGs vary in scope: while the UMLS is quite general, databases such as Orphanet focus specifically on rare diseases \citep{Weinreich2008orphanet}. These KGs can be generated in various ways. Some have used probabilistic models to extract data from patient notes \citep{Rotmensch2017learningKGs}, others use named entity recognition and other NLP techniques \citep{yu2022biosalgorithmicallygeneratedbiomedicalkg}, and many are built by reconciling a set of various sources \citep{Chandak2023primekg}. We use \citet{yu2022biosalgorithmicallygeneratedbiomedicalkg}'s biomedical informatics ontology system (BIOS) as ground truth to compare LLM--generated graphs to. 

\paragraph{The Relationship Between LLMs and Graphs} has been investigated in recent years. Causal graphs have found use in general medicine \citep{greenland2002causalmodellingmethodsepidemiology}, epidemiology \citep{greenland1999causaldiagramsforepidemiology}, and bioinformatics \citep{kleinberg2011causalinferenceforbiomedinformationcs}.  LLMs have been shown to find pairwise relationships \citep{kiciman2023causalreasoningandllms}, accurately determine edge direction, \citep{naik2023applyinglargelanguagemodels}, hypothesize missing variables \citep{sheth2024hypothesizing}, and be capable of generating small causal graphs with reasonable accuracy and efficiency \citep{long2024canllmsbuildcgs, jiralerspong2024efficientcgdiscoveryllms} and large KGs from texts \citep{ Hao2022-fj, Melnyk2022-aj, Zhang2023-cx}. LLMs have been also combined with statistical methods for generation \citep{ ban2023querytoolscausalarchitects,abdulaal2024causalmodellingagents,vashishtha2023causalinferenceusingllmguided}.
One reason why LLMs are so appealing for graph generation is that they are able to leverage metadata similarly to how a human expert would go about generating a causal graph \citep{kiciman2023causalreasoningandllms,abdulaal2024causalmodellingagents, choi2022lmpriorspretrainedlanguagemodels}. Augmenting LLM with KGs have been shown to improve task performance \citep{Jin2023-kd, Soman2023-zi, Jiang2023-zj}.
To the best of our knowledge, our work is the first work to build a complete graph from one given concept (going beyond pairwise comparison and partial graphs), which is used to evaluate LLMs for medical use.
 % While \cite{sheth2024hypothesizing} proposes to build up causal graphs, they however do from a partial causal graph. 
% . In many causal inference tasks, LLMs outperform vintage statistical methods \citep{kiciman2023causalreasoningandllms}. Recent work by  has further increased both the accuracy and efficiency of LLM CG generation. One reason why LLMs are so appealing for CG generation is that they are able to leverage metadata similarly to how a human expert would go about generating a CG \citep{kiciman2023causalreasoningandllms,abdulaal2024causalmodellingagents, choi2022lmpriorspretrainedlanguagemodels}. Some works have also leveraged both LLM-guided and statistical methods for the generation task \citep{ban2023querytoolscausalarchitects,abdulaal2024causalmodellingagents,vashishtha2023causalinferenceusingllmguided}. While \cite{sheth2024hypothesizing} proposes to build up causal graphs, they however do from a partial causal graph. 
% Our work is the first work to the best of our knowledge to build a complete graph from one given concept going beyond pairwise comparison and partial graphs.

% Knowledge graphs have been shown to improve LLM's task performance
% \ivaxi{NEED TO ADD res about KG+LLM there are almost none here}

% Although the vast majority---if not all---of these works generate directed acyclic graphs and strongly reside in the field of causal inference, observing their methods for generation and results greatly inspired our work. 

\section{Methodology}

\subsection{Preliminaries} 
A knowledge graph can be mathematically denoted as $ G = (V, E) $, where $ V $ defines a finite set of vertices or nodes and $ E $ is a set of ordered pairs of vertices. The vertices $ V $, are represented as $ \{v_1, v_2, \ldots, v_n\} $, with each $ v_i $ signifying a distinct entity or concept within the graph. The cardinality of $ V $, denoted $ |V| $, indicates the total number of entities represented in the graph. The edges are denoted as $ \{e_k=(v_i, v_j)\}_k$, where $v_i, v_j \in V$, $v_i \neq v_j$, and each $e_k$ represents a directed edge from node $ v_i $ to node $ v_j $. The presence of $e_k$ signifies a relationship or interaction between the entities represented by $ v_i $ and $ v_j $.

\subsection{MedG--KRP}
We introduce an algorithm based on the process of sequentially expanding from a given medical concept for the generation of biomedical KGs using LLMs. After LLMs generate graphs, a panel of medical students scores each graph based on accuracy and comprehensiveness. We then compare our LLM-generated KGs to the biomedical KG BIOS, computing precision and recall. 

\subsection{Generation Algorithm}

We divide our graph generation process into two primary stages: \textbf{node expansion} and \textbf{edge refinement}. In the first stage, nodes are recursively hypothesized by querying the LLM for relevant medical concepts, while the second stage involves validating and refining the edges between these nodes.

\subsubsection{Node Expansion}

\paragraph{}Our node expansion algorithm (Algorithm \ref{alg:expand}) aims to explore the causal relationships between medical concepts. The process begins with a root node $ r $, representing an initial medical concept, and recursively prompts an LLM for concepts that are either \textit{caused by} or \textit{cause} the root concept. The objective of this stage is to identify which medical concepts the LLM associates with $ r $, thereby capturing the model's understanding of the causal pathways surrounding a given medical condition or concept.

Formally, let $ r $ denote the root node, and $ x $ represent the current recursion depth. We expand the graph $ G $ by exploring both forward (causal) and backward (caused-by) relationships. The algorithm proceeds recursively, with each newly identified node being further expanded to find related concepts.

To prevent unbounded expansion and ensure the graphs remain interpretable, we impose a maximum recursion depth of 2. Additionally, to maintain legibility and minimize the risk of hallucination, we limit the LLM to returning at most $ n_{\max} = 3 $ concepts in response to each query. Importantly, there is no lower bound on the number of concepts an LLM may return; the LLM can indicate that there are no concepts either causing or caused by a given node, which helps maintain the algorithm's reliability and reduces over-expansion.

\begin{algorithm}[htbp]
\floatconts
  {alg:expand}%
  {\caption{Recursive Node Exploration}}
{
{
\KwIn{$r$, a root concept node,

\qquad\quad\,\,$n_{\max}=3$, maximum concepts for one response,

\qquad\quad\,\,$x_{\max}$, maximum recursion depth.

\qquad\quad\,\,\textsc{Expand-o($r, n_{max}$)}, function to prompt LLM to expand in the outwards (caused by) direction relative to $r$, returning $n_{max}$ concepts, following prompt \ref{exp_left},

\qquad\quad\,\,\textsc{Expand-i($r, n_{max}$)}, function to prompt LLM to expand in the inwards (causing) direction relative to $r$, returning $n_{max}$ concepts, following prompt \ref{exp_right},

\qquad\quad\,\,$d$, direction (i.e. \emph{causing} or \emph{caused by}).

}
\KwOut{$G=(V,E)$. A KG with nodes explored}
\textbf{Initialize}: $V = \{r\}$, $E = \varnothing$ 
}
\BlankLine
\BlankLine
$G \gets$ \textsc{nodefind($r, 0, ``caused\ by", G$)}

$G \gets$ \textsc{nodefind($r, 0, ``causing", G$)}
\BlankLine
\textsc{nodefind($r, x, d, G$):} 
\If{$x > x_{\max}$}{\KwRet{$G$}}
\If{$d$ = \textrm{``caused by"}}{
 \tcc{ask LLM for concepts caused by r}
$V_{\textrm{caused\_by}}:=\{v_{1}, \ldots ,v_{n}\} =$ \textsc{Expand-o($r, n_{max}$)}

$V \gets V \cup V_{\textrm{caused\_by}}$

$E \gets E \cup \{(r, v) : v \in V_{\textrm{caused\_by}}\}$

\For{$v \in V_{\textrm{caused\_by}}$}{
    \textsc{nodefind($v, x + 1, ``\emph{caused by}")$}
    \textsc{nodefind($v, x + 1, ``\emph{causing}")$}
}
}
\If{$d$ = \textrm{``causing"}}{
 \tcc{ask LLM for concepts causing r}
$V_{\textrm{causing}}:=\{v_{1}, \ldots ,v_{m}\} =$ \textsc{Expand-i($r, n_{max}$)}

    $V \gets V \cup V_{\textrm{causing}}$
    
    $E \gets E \cup \{(v, r) : v \in V_{\textrm{causing}}\}$
    
    \For{$v \in V_{\textrm{causing}}$}{
        \textsc{nodefind($v, x + 1, ``\emph{caused by}")$}
        \textsc{nodefind($v, x + 1, ``\emph{causing}")$}
    }

}
}
\end{algorithm}
\vspace{-3mm}
\subsubsection{Edge refinement} 

In the second stage (Algorithm \ref{alg:check}), we perform an exhaustive check for additional causal connections that the LLM may infer should exist between the concepts already present in the graph. This step is crucial for ensuring the completeness of the knowledge graph by identifying all potential relationships between nodes. It is quite possible that, after the node expansion algorithm has been run, there are edges that are not yet present in the current graph but ideally would be. 

% Given a pair of nodes \(a\) and \(b\), we query the LLM to determine if a directed edge \((a, b)\) exists. Similarly, the reverse direction \((b, a)\) is also queried, treating these two directions as distinct. This allows for the possibility of bidirectional edges, representing mutual causality or interdependence in medical contexts.

% We opt to query each direction separately, rather than including all possible edge directions in a single query, in order to reduce the cognitive load on the LLM. By isolating each query to a single direction, we hypothesize that the LLM can provide more accurate predictions regarding the presence of specific edges. -> in the abalation results
Let \(G\) denote the graph of concepts obtained after the expansion stage. For each pair of distinct nodes \(v_i, v_j \in V\), where \(v_i \neq v_j\), we query the LLM for the existence of a directed edge \((v_i, v_j)\). If the LLM confirms that such an edge should exist, it is added to the graph. This process is repeated for every pair of nodes and for both directions, ensuring that the graph captures all potential causal relationships based on the LLM's understanding.

We opt to query each direction separately, rather than including all possible edge directions in a single query, in order to reduce the cognitive load on the LLM. By isolating each query to a single direction, we hypothesize that the LLM can provide more accurate predictions regarding the presence of specific edges. This method also allows for the possibility of bidirectional edges, representing mutual causality or interdependence in medical contexts.

\begin{algorithm}[htbp]
\floatconts
{alg:check}%
{\caption{Edge Refinement}}  
{
{
\KwIn{$G=(V,E)$, an incomplete generated KG,

\qquad\quad\,\,\textsc{Prompt($v, u$)}, function to ask LLM if an edge between two nodes exists or not, following prompt \ref{edge_ref}.

}
\KwOut{$G' = (V, E'),$ a KG with residual edges found defined over the nodes in G.}

\textbf{Initialize:} $E' = E$ \\ 
}

\tcc{loop through all node pairs, excluding pairs containing a node and itself}
\ForEach{$v \in V$}{ 
    \ForEach{$u \in V \setminus    \{v\}$}
    {
        % \If{$a \neq b$}{
            % Query LLM if edge \((a, b)\) exists\\
            % \If{LLM confirms \((a, b)\)}{
             \,\,\tcc{ask LLM if an edge should be created between the nodes} \If{\textsc{Prompt}\((v, u)\)}{
                $E' \gets E' \cup \{(v, u)\}$
            }
        %}
    }
}

}
\end{algorithm}

\section{Experimental Setup}

\subsection{Concept Selection} 
We selected twenty conditions from various sub-disciplines of medicine to act as the root nodes for our graphs. We chose a list of conditions that would vary vastly in prevalence and level of study. We include both conditions with clear causal pathways and unclear ones. A full list of root concepts, verified by a board--certified physician, can be found in Table \ref{tab:avgs}.

\subsection{Models} We tested our benchmark on diverse models--the propriety GPT-4 model~\citep{OpenAI2023-hq}, open source Llama3-70b~\citep{Dubey2024-qg}, and finally the current state-of-art medical model PalmyraMed-70b~\citep{Palmyra-Med-70B}. PalmyraMed--70b is a Llama base model fine-tuned for medical usage which displays very good performance on medical LLM benchmarks. We aimed to compare the performance of a medical finetune model in comparison to its base model counterpart.

\subsection{Hyperparameters} We run Algorithm \ref{alg:expand} in both directions for the graph $G$, exploring concepts that either cause or are caused by the root medical concept $r$. Starting with iteration $x = 1$, we limit the maximum depth of recursion $depth_{max}$ to 2, meaning the \textsc{Nodefind} function calls itself only once. After completing Algorithm \ref{alg:expand}, the graph $G$ will contain all relevant nodes. To identify additional directed edges between the concepts in $G$, we then execute Algorithm \ref{alg:check}.

All models are evaluated with a temperature setting of 0.05 and a top\_p value of 1.0. The low temperature ensures that results primarily reflect the models' reasoning abilities, enhancing reproducibility. We do not set the temperature to 0.0 to allow for slight variations in responses during re-prompting, should issues arise if a model doesn't format its answers as we request.

\subsection{Prompting}
In this paper, we aim to evaluate the ability of LLMs to hypothesize knowledge graphs using their zero-shot prompting abilities. Three main prompts were used, one system prompt and one general prompt for each algorithm. 

\paragraph{System Prompt}
The system prompt (see Appendix \ref{apd:second}, section \ref{sys_prompt}) is designed to enhance the LLM’s reasoning ability by focusing on distinguishing direct and indirect causality—an area where LLMs often struggle. To improve response quality, we instruct the model to employ counterfactual reasoning, asking it to evaluate causal relationships by considering hypothetical scenarios.

\paragraph{Expansion Prompts}
Two expansion prompts (see Appendix \ref{apd:second}, Section \ref{exp_left}, Section \ref{exp_right}) are used in Algorithm \ref{alg:expand} to discover concepts related to the root node, one for \say{causes} and one for \say{caused by.} Similar to the system prompt, we emphasize counterfactual reasoning to improve accuracy. We employ a zero-shot chain-of-thought (CoT) approach, following \citet{kojima2023llmszeroshot}, which enhances performance in medical QA tasks and pairwise edge-checking. The current graph state is passed into each prompt to maintain context during the expansion process.

\paragraph{Edge Check Prompt}
The edge check prompt (see Appendix \ref{apd:second}, Section \ref{edge_ref}), as used in Algorithm \ref{alg:check}, queries the LLM to determine if a directed causal relationship exists between two medical concepts. Like the system and expansion prompts, we emphasize distinguishing direct from indirect causality using counterfactual reasoning. To isolate the causal connection, the prompt assumes no external risk factors are influencing the relationship, ensuring that the LLM focuses on the specific medical concepts being tested.

\subsection{Metrics}

\paragraph{Human Evaluation: Graph Accuracy and Comprehensiveness} We enlisted a panel of medical students to manually comment on and score all generated graphs in terms of accuracy and comprehensiveness. We define accuracy as medical correctness of all concepts, relationships, and implied causal pathways in a given graph. We define a graph for a given disease to be completely comprehensive if a human reviewer believes that the graph covers all medical concepts that should be covered for a proper understanding of the given disease. Thus, graphs with many missing nodes that would be present in a hypothetical ideal graph representing current medical understanding would have low comprehensiveness scores. Accuracy was scored from a scale of 1-4; [Completely accurate (4), Mostly Accurate (3), Inaccurate (2), Completely inaccurate (1)]. Comprehensiveness was scored similarly, on the scale [Completely Comprehensive (4), Mostly Comprehensive (3), Poorly Comprehensive (2), Not At All Comprehensive (1)]. Three reviewers scored each graph. We report all reviewer scores and an average thereof for each graph.

\paragraph{Ground Truth Comparison} Generated graphs were also compared to the Biomedical Informatics Ontology System (BIOS). BIOS is a large knowledge graph composed of numerous sources and containing hundreds of thousands of nodes and edges. We chose it because BIOS appeared more complete and suitable for our use case than other biomedical KGs. We calculated the precision and recall of generated edges using \autoref{alg:prec_recall}. For each generated graph, we iterate through all edges and check if there is a path of length less than or equal to 7 between the two corresponding concepts in the ground truth. This means that, for a given edge, the number of intermediary nodes in the ground truth between the two nodes that constitute the edge must be less than or equal to five. If a path in the ground truth satisfies this condition, it is marked as a hit. Otherwise, it is marked as a miss. The intent of checking for paths instead of a direct edge is to avoid the case where an edge in a generated graph would be deemed as medically correct but have intermediary concepts in the ground truth.

% \begin{algorithm}[htbp]
% \floatconts
% {alg:prec_recall}%
% {\caption{Precision and recall}}  
% {
% \KwIn{$G=(V,E)$, the generated KG,

% \qquad\quad\,\,$G'=(V',E')$, the BIOS KG,

% \qquad\quad\,\,\textsc{Path($v_{1}, v_{2}$)}, function to return the shortest path between two nodes,

% \qquad\quad\,\,d = 5, the maximum number of intermediate nodes in a path.
% }
% \KwOut{precision and recall over edges.}

% $V'' = V \cap V'$

% $E'' = E \cap E'$

% $E_{rel} = \{(v'_{i}, v'_{j}) \in E' : v'_{i} \in V \lor v'_{j} \in V\}$ \tcc{edges in BIOS relevant to the generated KG}

% $G''=(V'', E'')$

% \ForEach{$(v_{i},v_{j}) \in E$}{
%     \If{$\{(v_{i},v_{j})\}\subseteq E''$ and \\
%     \tcc{a path is found that does not exceed our threshold for intermediate nodes...}
%     $\qquad\!\!\!\exists \textsc{Path}(v_{i},v_{j})\subseteq E''$ s.t. $| \textsc{Path}(v_{i},v_{j})| - 2\leq d$}{
%      $n_{\mathrm{hit}} \gets n_{\mathrm{hit}} + 1$
%     }
% }
% Precision = $n_{\mathrm{hit}}/|E|$ 

% Recall = $n_{\mathrm{hit}}/{|E_{rel}|}$

% }
% \end{algorithm}

\begin{algorithm}[htbp]
\floatconts
{alg:prec_recall}%
{\caption{Precision and recall}}  
{
\KwIn{$G=(V_g,E_g)$, the generated KG,

\qquad\quad\,\,$B = (V_b,E_b)$, the BIOS KG,

\qquad\quad\,\,\textsc{Path$_B$($v, u$)}, function to return the shortest path between node $v$ and node $u$ in graph $B$,

\qquad\quad\,\,d = 7, the threshold for maximum path length (five intermediary nodes). 
}
\KwOut{precision and recall over edges.}

% $V_o = V_g \cap V_b$ \tcc{overlapped vertices}
% $E_o = E_g \cap E_b$ \tcc{overlapped edges}

\tcc{edges in the generated KG defined over nodes in the BIOS graph}

$E_{gb} = \{(v_{g}, u_{g}) \in E_{g} : v_{g} \in V_{b} \land u_{g} \in V_{b}\}$

\tcc{for each edge in the generated KG defined over nodes in BIOS, check if a short path exists in BIOS}
\ForEach{$(v_g,u_g) \in E_{gb}$}{
    \If{%$\{(v_{i},v_{j})\}\subseteq E''$ \text{and} \\
    $\exists P:=\textsc{Path}_B(v_g,u_g)$ \text{s.t.} $|P| \leq d$}{
     $n_{\mathrm{hit}} \gets n_{\mathrm{hit}} + 1$
    }
}
Precision = $n_{\mathrm{hit}}/|E_g|$ 

$E_{rel} = \{(v_{b}, u_{b}) \in E_b : v_{b} \in V_g \lor u_{b} \in V_g\}$ \tcc{edges in BIOS relevant to the generated KG}

Recall = $n_{\mathrm{hit}}/{|E_{rel}|}$

}
\end{algorithm}
% In order to map nodes from an LLM generated graph to their corresponding nodes in BIOS, 
\subparagraph{Mapping node names} We directly match BIOS graph and LLM-generated graphs by building a vector database with embeddings of all BIOS parent concepts using the sentence transformers model e5--base--v2 ~\citep{wang2022text}. Our decision to use e5 as opposed to a specialist medical model was motivated by our belief that nodes out of the scope of medicine, which a medical model could have trouble handling, could be present in LLM--generated graphs. We retrieved the five nearest neighbors in the vector database, then prompted GPT-4 as to which (if any) names of the nearest neighbor nodes matched a given node in an LLM-generated graph in meaning. The intention of this process is to handle the rare case in which the nearest neighbor in the vector database isn't identical to the lookup vector in meaning. The prompt used can be found in Appendix \ref{apd:second}, Section \ref{nn_prompt}. We iterated through all generated nodes and created a JSON file with every LLM generated node and its BIOS counterpart. If a node had no counterpart, we simply set its value to \say{none}. 

% \subparagraph{Relatedness} For an LLM generated graph, we iterate through all edges and check if there is a path between the two corresponding concepts in the ground truth which is of less length than threshold \(t\). In our case, we define \(t = 5\). If a path in the ground truth satisfies this condition, it is marked as a hit. Otherwise, it is marked as a miss. The intent of checking for paths in the ground truth instead of a direct edge between two concepts is to avoid the case where an edge in a generated graph would be deemed as medically correct but have intermediary concepts in the ground truth. 

%\\\qquad\quad\,\,$G''=\{V_G \bigcap V_{G'}, \{(u,v)\in E_G\}\}$, the BIOS graph.

% \vspace{-3mm}

% \subparagraph{Calculating precision and recall} Let \(G=(V_G,E_G)\) denote an LLM--generated graph and \(G'=(V_G',E_G')\) denote the subgraph of BIOS defined over the $V_{G'}$. For every node pair \(i, j \in E_{G}\), retrieve the two corresponding nodes, \(i^\prime, j^\prime\) in \(V_{G'}\). If a path of length no longer than the threshold \((t=5\)) exists between \(i^\prime, j^\prime\) in \(G^\prime\), then the edge \(i, j\) is said to be correct and  $n_{\mathrm{hit}}$ is incremented. Otherwise, the edge is false. All edges of nodes that did not map to BIOS are considered false. Thus, our precision and recall scores are given as: 

% \begin{align*}
% Precision = \frac{n_{\mathrm{hit}}}{|E_{G}|} \\ 
% Recall = \frac{n_{\mathrm{hit}}}{|E_{G^\prime}|}
% \end{align*}

\subparagraph{Edge types} While all edges in \emph{MedG--KRP} generated graphs specify ``cause", edges in BIOS contain many specific relationship labels. We do not recognize any edges in BIOS labeled as ``is a" or ``reverse is a" because both are used only for subclasses or superclasses of a given concept and because of performance constraints. A consequence of this is, due to BIOS's incompleteness, some nodes are not reachable. The remaining edges are a mix of bidirectional and directional edges, so we interpret all edges as bidirectional for the sake of consistency, and due to the fact that the directions of all edges in BIOS are implied by their labels, rather than explicitly stated.  

\section{Results}

% \begin{table*}[hbtp]
% \floatconts
%   {tab:avgavg}
%   {\caption{Average of all Reviewer Scores per Model}}
%   {\begin{tabular}{lccc}
%   \toprule
%   \bfseries  & \bfseries Llama3--70b & \bfseries PalmyraMed & \bfseries GPT-4\\
%   \midrule
%   Acc. & 3.2833 & 3.1333 & 3.3667\\
%   Comp. & 3 & 2.9667 & 3.225\\
%   \bottomrule
%   \end{tabular}}
% \end{table*}

\begin{table*}[htb!]
\small
\floatconts
  {tab:avgs}
  {\caption{Mean Reviewer Scores (from 1--4) per Graph per Model}}
  {\begin{tabular}{lcccccc|cc}   
    \toprule
     & \multicolumn{2}{c}{\textbf{Llama3--70b}} & \multicolumn{2}{c}{\textbf{PalmyraMed}} & \multicolumn{2}{c}{\textbf{GPT-4}} & \multicolumn{2}{c}{\textbf{Average}}
    \\\cmidrule(lr){2-3}\cmidrule(lr){4-5}\cmidrule(lr){6-7}\cmidrule(lr){8-9}
         & Acc. & Comp. & Acc. & Comp. & Acc. & Comp. & Acc. & Comp.\\\midrule
    Acute flaccid myelitis                    & $\underset{\pm0.33}{\textbf{3.67}}$ & $\underset{\pm1.00}{3.00}$ & $\underset{\pm0.33}{2.67}$ & $\underset{\pm0.33}{3.33}$ & $\underset{\pm0.33}{3.33}$ & $\underset{\pm0.33}{\textbf{3.67}}$ & 3.22 & 3.33 \\
    Arthritis                                 & $\underset{\pm0.00}{3.00}$ & $\underset{\pm0.33}{\textbf{3.33}}$ & $\underset{\pm0.33}{2.67}$ & $\underset{\pm1.00}{3.00}$ & $\underset{\pm0.33}{\textbf{3.33}}$ & $\underset{\pm.33}{\textbf{3.33}}$ & 3.00 & 3.22 \\
    Asthma                                    & $\underset{\pm0.33}{\textbf{3.33}}$ & $\underset{\pm1.00}{3.00}$ & $\underset{\pm0.33}{2.33}$ & $\underset{\pm1.00}{3.00}$ & $\underset{\pm0.33}{\textbf{3.33}}$ & $\underset{\pm0.00}{\textbf{4.00}}$ & 3.00 & 3.33 \\
    Creutzfeldt--Jakob disease                 & $\underset{\pm1.33}{2.67}$ & $\underset{\pm0.33}{2.67}$ & $\underset{\pm0.33}{2.67}$ & $\underset{\pm0.33}{\textbf{3.33}}$ & $\underset{\pm0.33}{\textbf{3.33}}$ & $\underset{\pm0.00}{3.00}$ & 2.89 & 3.00 \\
    Dementia                                  & $\underset{\pm0.33}{3.33}$ & $\underset{\pm1.33}{3.33}$ & $\underset{\pm0.33}{3.33}$ & $\underset{\pm0.33}{2.33}$ & $\underset{\pm0.00}{\textbf{4.00}}$ & $\underset{\pm0.33}{\textbf{3.67}}$ & 3.56 & 3.11 \\
    Diabetes Mellitus                         & $\underset{\pm0.00}{\textbf{4.00}}$ & $\underset{\pm0.33}{\textbf{3.67}}$ & $\underset{\pm0.33}{\textbf{3.67}}$ & $\underset{\pm1.00}{3.00}$ & $\underset{\pm0.00}{3.00}$ & $\underset{\pm0.58}{2.83}$ & 3.56 & 3.17 \\
    Esophageal achalasia                      & $\underset{\pm0.00}{3.00}$ & $\underset{\pm0.00}{3.00}$ & $\underset{\pm2.33}{2.67}$ & $\underset{\pm1.00}{3.00}$ & $\underset{\pm0.33}{\textbf{3.67}}$ & $\underset{\pm1.00}{3.00}$ & 3.11 & 3.00 \\
    Glioblastoma                              & $\underset{\pm1.33}{2.67}$ & $\underset{\pm1.00}{2.00}$ & $\underset{\pm0.00}{3.00}$ & $\underset{\pm1.00}{3.00}$ & $\underset{\pm0.33}{2.67}$ & $\underset{\pm1.33}{\textbf{3.33}}$ & 2.78 & 2.78 \\
    HIV                                       & $\underset{\pm0.33}{3.33}$ & $\underset{\pm0.33}{2.33}$ & $\underset{\pm0.33}{3.33}$ & $\underset{\pm1.00}{\textbf{3.00}}$ & $\underset{\pm0.33}{3.33}$ & $\underset{\pm2.33}{2.33}$ & 3.33 & 2.56 \\
    Hyperparathyroidism                       & $\underset{\pm0.33}{3.33}$ & $\underset{\pm0.33}{2.67}$ & $\underset{\pm0.00}{3.00}$ & $\underset{\pm1.00}{\textbf{3.00}}$ & $\underset{\pm0.00}{\textbf{4.00}}$ & $\underset{\pm1.00}{\textbf{3.00}}$ & 3.44 & 2.89 \\
    Ischemic Stroke                           & $\underset{\pm0.33}{3.67}$ & $\underset{\pm0.00}{\textbf{4.00}}$ & $\underset{\pm0.00}{\textbf{4.00}}$ & $\underset{\pm0.00}{3.00}$ & $\underset{\pm0.33}{3.67}$ & $\underset{\pm2.33}{2.67}$ & 3.78 & 3.22 \\
    Lung Cancer                               & $\underset{\pm0.33}{\textbf{3.67}}$ & $\underset{\pm0.33}{2.33}$ & $\underset{\pm0.33}{\textbf{3.67}}$ & $\underset{\pm0.33}{2.67}$ & $\underset{\pm0.00}{3.00}$ & $\underset{\pm0.00}{\textbf{4.00}}$ & 3.44 & 3.00 \\
    Malignant neoplasms of liver              & $\underset{\pm0.33}{3.67}$ & $\underset{\pm0.00}{\textbf{3.33}}$ & $\underset{\pm0.33}{3.67}$ & $\underset{\pm1.00}{\textbf{3.33}}$ & $\underset{\pm0.00}{\textbf{4.00}}$ & $\underset{\pm0.33}{2.67}$ & \textbf{3.78} & 2.89 \\
    Myocardial infarction                     & $\underset{\pm0.33}{3.33}$ & $\underset{\pm2.33}{2.67}$ & $\underset{\pm0.00}{\textbf{4.00}}$ & $\underset{\pm0.00}{\textbf{3.00}}$ & $\underset{\pm0.33}{3.33}$ & $\underset{\pm1.00}{\textbf{3.00}}$ & 3.56 & 2.89 \\
    Myocarditis                               & $\underset{\pm0.33}{\textbf{3.67}}$ & $\underset{\pm1.00}{\textbf{3.00}}$ & $\underset{\pm0.33}{3.33}$ & $\underset{\pm0.33}{2.67}$ & $\underset{\pm1.00}{3.00}$ & $\underset{\pm1.00}{\textbf{3.00}}$ & 3.33 & 2.89 \\
    Parkinson’s disease                    & $\underset{\pm0.00}{3.00}$ & $\underset{\pm0.00}{\textbf{3.00}}$ & $\underset{\pm0.33}{\textbf{3.33}}$ & $\underset{\pm0.33}{2.33}$ & $\underset{\pm1.00}{3.00}$ & $\underset{\pm1.00}{\textbf{3.00}}$ & 3.11 & 2.78 \\
    Renal artery stenosis                     & $\underset{\pm0.33}{3.33}$ & $\underset{\pm0.33}{\textbf{3.67}}$ & $\underset{\pm0.33}{3.67}$ & $\underset{\pm0.33}{\textbf{3.67}}$ & $\underset{\pm0.00}{\textbf{4.00}}$ & $\underset{\pm0.33}{3.33}$ & 3.67 & \textbf{3.56} \\
    SARS-CoV-2                                & $\underset{\pm1.00}{3.00}$ & $\underset{\pm0.33}{3.33}$ & $\underset{\pm1.00}{3.00}$ & $\underset{\pm0.33}{\textbf{3.67}}$ & $\underset{\pm0.33}{\textbf{3.33}}$ & $\underset{\pm0.33}{\textbf{3.67}}$ & 3.11 & \textbf{3.56} \\
    Spontaneous coronary artery dissection    & $\underset{\pm0.00}{\textbf{3.00}}$ & $\underset{\pm3.00}{3.00}$ & $\underset{\pm1.00}{2.00}$ & $\underset{\pm0.33}{2.67}$ & $\underset{\pm0.00}{\textbf{3.00}}$ & $\underset{\pm0.33}{\textbf{3.67}}$ & 2.67 & 3.11 \\
    Ulcerative colitis                        & $\underset{\pm1.00}{\textbf{3.00}}$ & $\underset{\pm1.00}{3.00}$ & $\underset{\pm0.33}{2.67}$ & $\underset{\pm0.33}{2.67}$ & $\underset{\pm1.00}{\textbf{3.00}}$ & $\underset{\pm1.33}{\textbf{3.33}}$ & 2.89 & 3.00 \\

    \midrule
    Average Score  & 3.28 & 3.00 & 3.13 & 2.97 & \textbf{3.37} &\textbf{3.23}  \\
    Average Variance & 0.42 & 0.72 & {0.43} & 0.57 & 0.32 & {0.76} \\
    \bottomrule
    \end{tabular}}
\end{table*}

\begin{table}[hbt!]
\footnotesize
\floatconts
  {tab:precisionrecall}
  {\caption{Average Precision and Recall per Model}}
  {\begin{tabular}{lcccccc}
  \toprule
   & \multicolumn{2}{c}{\scriptsize{\textbf{Llama3--70b}}} & \multicolumn{2}{c}{\scriptsize{\textbf{PalmyraMed}}} & \multicolumn{2}{c}{\scriptsize{\textbf{GPT-4}}}
  \\\cmidrule(lr){2-3}\cmidrule(lr){4-5}\cmidrule(lr){6-7}
       & Pre. & Rec. & Pre. & Rec. & Pre. & Rec. \\\midrule
  Mean & .201   & .012   & \textbf{.243}   & \textbf{.033}   & .163   & .011     \\
  Min. & .000   & .000   & \textbf{.026}   & \textbf{.004}   & .018   & .003     \\
  Max. & .486   & .034   & \textbf{.527}   & \textbf{.393}   & .359   & .031     \\
  SD   & .123   & .008   & \textbf{.150}   & \textbf{.085}   & .106   & .007     \\
  \bottomrule
  \end{tabular}}
\end{table}

% \begin{table}[hbtp]
% \footnotesize
% \floatconts
%   {tab:nodesedges}
%   {\caption{Graph Node and Edge Counts per Model}}
%   {\begin{tabular}{lcccccc}
%   \toprule
%    & \multicolumn{2}{c}{\textbf{Llama3--70b}} & \multicolumn{2}{c}{\textbf{PalmyraMed}} & \multicolumn{2}{c}{\textbf{GPT-4}}
%   \\\cmidrule(lr){2-3}\cmidrule(lr){4-5}\cmidrule(lr){6-7}
%        & nodes & edges & nodes & edges & nodes & edges \\\midrule
%   Mean & 31.5   & 78.25   & 27      & 72.05    & 33.4    & 85.65   \\
%   Min. & 26     & 45      & 17      & 49       & 25      & 39.     \\
%   Max. & 36     & 120     & 34      & 172      & 38      & 146     \\
%   SD   & 3.32 & 17.93 & 4.09  & 27.49  & 3.87  & 29.46 \\
%   \bottomrule
%   \end{tabular}}
% \end{table}

\subsection{Overview}
\paragraph{} We generated sixty graphs across three models for twenty different conditions from various fields of medicine. We observe that \textbf{all LLMs perform generally well in terms of average reviewer scores} (see Table \ref{tab:avgs}). GPT-4 displays the strongest performance in the human review, while PalmyraMed displays the weakest. Human reviewers generally found that PalmyraMed's graphs are more specific than those generated by Llama3--70b and GPT-4. Even for the same model, generated graphs have a wide variety of density values, reciprocity values, and simple cycle counts. 

\subsection{Human Evaluation}
\paragraph{} Accuracy, as rated by human reviewers, is generally strong, with all averages of all reviewer scores for each model being between 3 and 4, ``mostly accurate" and ``completely accurate" (see Table \ref{tab:avgs}). Comprehensiveness scores range from just under 3 to 4. We attribute comprehensiveness scores consistently being lower than accuracy scores to us limiting responses and recursion depth in our recursive node exploration algorithm (see Algorithm \ref{alg:expand}).

GPT-4 performed best in accuracy, with an average accuracy score across all graphs of 3.37 (see Table \ref{tab:avgs}. Llama3 was close behind with an average accuracy of 3.28 and PalmyraMed displayed the worst performance with a score of 3.13. Both Llama3 and PalmyraMed performed similarly in comprehensiveness, with average comprehensiveness scores across all graphs of 3.00 and 2.97 (see Table \ref{tab:avgs}. GPT-4 displayed the best comprehensiveness, with a score of 3.23--a significantly stronger performance than all other models.  

In their comments, reviewers mentioned that PalmyraMed's graphs were generally more specific than those of GPT-4 and Llama3--70b. We speculate this to be a result of PalmyraMed being aligned for medical usage.

%: Being trained on more domain specific data, we hypothesize that PalmyraMed contain more medical knowledge whereas  generalist models like GPT-4 and Llama3--70b contain more public knowledge.

Llama3--70b having weaker overall performance than GPT-4 follows its generally weaker performance on traditional QA benchmarks. PalmyraMed, however, has been shown to have better average performance on QA benchmarks than GPT-4, yet it performed worse overall on our benchmark. Reviewers noticed that \textit{PalmyraMed appeared much more prone to hallucination} than other models, with it naming multiple graph nodes ``myra-med'' or ``PalmyraMed'', and having trouble with instruction following.  

\subsection{Ground Truth Comparison}
We observe notable results in the ground truth comparison metric, where models demonstrated behavior nearly opposite to that observed in human evaluations (see tables \ref{tab:precisionrecall}, \ref{tab:avgs}). PalmyraMed performed exceptionally well, with the highest precision and recall scores across the board. In particular, PalmyraMed displayed more than three times the average recall score of GPT-4, which displayed the worst performance. Interestingly enough, Llama3--70b, which is usually surpassed by GPT-4 on almost all major QA benchmarks, outperformed GPT-4 in both precision and recall in objective evaluation. 

\subsection{Graph Attributes}
\paragraph{} Overall graph node and edge counts (see Table \ref{tab:nodesedges}) varied between models and between graphs. \textit{PalmyraMed was generally the most conservative when creating nodes and edges} while GPT-4 was the least, possibly contributing to PalmyraMed's low comprehensiveness. 

We observe an inverse relationship between performance based on reviewer scores and average reciprocity, density, and simple cycle count across all graphs for a given model. GPT-4 displayed the highest performance and the lowest reciprocity, density, and simple cycle counts, while PalmyraMed displayed the highest. Llama3--70b's values for these three metrics were in--between those of the other models. 

Simple cycle counts for graphs for a given model varied widely. Each model consistently displayed one or two graphs which were outliers in terms of simply cycle count. PalmyraMed had the most extreme outlier, with its graph for ``Malignant neoplasms of liver'' containing more than one million cycles, while all other graphs contained under 2500 and all graphs in the bottom 75th percentile contained less than one hundred. Llama3's graph for ``creutzfeldt--jakob disease'' has a simple cycle count of greater than 3500 and all other graphs display cycle counts less than 225. GPT-4 displays the most reasonable cycle counts out of all models in its generated graphs, with one outlier of 340 and a bottom 50th percentile of less than or equal to eight cycles.

\subsection{Direct and Indirect Causality}
\paragraph{} Reviewers found that PalmyraMed often had difficulty distinguishing direct and indirect causality. Some reviewers mentioned that PalmyraMed often listed nodes as \say{causes} that would be much more appropriately labeled as \say{risk factors} GPT-4, on the other hand, was observed by reviewers to display the strongest ability to distinguish between directy and indirect causality, an ability crucial in medicine.

\section{Discussion}
\subsection{Conclusions}
\paragraph{} Our algorithm, \emph{MedG--KRP}, is able to generate KGs representing the medical reasoning abilities of LLMs. Coupling \emph{MedG--KRP} with human reviewers allowed insights into model behavior that were not covered by traditional QA benchmarks. We found that PalmyraMed was generally more specific in its reasoning, but also had a weaker understanding of the differences between direct and indirect causality, while GPT-4 covered more broad concepts and was often able to correctly determine between direct and indirect causes of concepts. 
%The ability of our method to find these crucial insights acts as a testament to its value. 

Although PalmyraMed displayed worse performance in our human review compared to other models, its KG---while flawed---was more specific than that of other models. This is supported by PalmyraMed's exceptionally high recall on our KG comparison task. We hypothesize that PalmyraMed, as a medical model, was trained on similar sources to which BIOS was constructed from than other LLMs we tested. This would lead to more frequent matches between nodes generated by PalmyraMed and BIOS nodes. Since nodes without mappings would have all adjacent edges generated counted as misses, it follows that a model that produced nodes more similar to those in BIOS would have much higher recall.

Clinicians may see PalmyraMed's specificity as a desirable trait. GPT-4 and Llama3--70b using more vague terms may signal that they are more influenced by public knowledge than by clinical knowledge since they are generalist models. It is worth noting that models were asked to be particularly specific and to stay to only medical---as opposed to colloquial---terminology. A human doctor whose reasoning was based on public discourse over medical understanding would not be trusted. Likewise, although expected, generalist LLMs having less specific KGs may suggest value in aligning models for clinical use. We wish to once again stress that the ability to find these observations is possible with our method, but not necessarily covered by traditional QA benchmarks. 

\subsection{Future Work}
Given that reviewers observe generalist models have a better causal reasoning ability compared to the medical model we tested but are lacking in domain specificity, the question of how we can build models that display both of these abilities naturally arises. Future works may seek to supplement the training corpora of traditional medical models with information on causal inference and causal reasoning to improve models' medical understanding and viability for real--world application.

We also believe that attempting to explore LLMs' internal KGs that are unrelated to medicine may yield interesting results. The topics of KGs could be from any field, and seeing how LLMs' reasoning changes when encountering vastly different subjects could give deeper insight into LLMs' behaviors. 

Using \emph{MedG--KRP} or a similar algorithm as a prompting technique may also be possible. An LLM could generate a reasoning graph then be prompted to make inferences or answer questions given the graph it produced like CoT prompting. 

Other pathways that may be worthwhile to explore include, in no specific order: exploring the effect of an LLM's training data on its reasoning KGs, using KG generation to determine the effect (if any) of pretraining data order on LLM behavior, revising the \emph{MedG--KRP} algorithm or developing new algorithms to efficiently use directly prompted LLMs for biomedical KG generation or repair, and building very large reasoning KGs with LLMs to probe behavior at a larger scale and how and when LLMs connect interdisciplinary or 
seemingly unrelated concepts. 
\bibliography{jmlr-sample}
\newpage
\appendix
\section{Limitations}\label{apd:first}
\paragraph{} While we test our method on a diverse set of models, others may show very different behavior. Although we aimed to make our list of diseases used for graph generation broad, it is by no means comprehensive. Due to its time complexity, our approach is also only suitable for the generation of small graphs---sufficient for benchmarking purposes but not for full generation of KGs. Our human review is subjective, and only three reviewers go over a given graph. In addition, we found there was often high variance in reviewer opinions. The knowledge graph we use as ground truth, BIOS, may also be quite incomplete. We also only test using one knowledge graph as ground truth. We believe that, in the case that a human expert scored every graph edge, precision values would be greater than the ones which we report. 

\section{Prompts}\label{apd:second}
\subsection{System Prompt}
\label{sys_prompt}
\begin{verbatim}
You are a helpful assistant for causal 
inference and causal reasoning about medical 
questions. You are always specific in your 
answers. You always format your answers 
consistently and name all medical terms in 
the correct and accepted medical lexicon. 
You understand the differences between 
direct and indirect causality and 
acknowledge these differences when 
formulating an answer. You utilize a 
counterfactual model of causal inference 
when formulating a response. 
\end{verbatim}

\subsection{Left Expansion Prompt}
\label{exp_left}
\begin{verbatim}
A directed knowledge graph that you 
generated is surrounded in XML tags and 
provided below. This directed knowledge 
graph is formatted as a list of edges like 
so: ['a causes b', 'b causes c', etc]. The 
knowledge graph you generated is as follows: 

<Begin Knowledge Graph> 
{edges:} 
</End Knowledge Graph> 

Given the directed knowledge graph above 
that you generated, up to {n_max:} factors 
that directly cause {concept:}. These factors 
do not need to be in the knowledge graph 
above, but can be. If a factor you answer with 
is in the knowledge graph above, in your 
response, name it exactly as it is named in 
the graph above. Do not answer with any 
factors that only indirectly cause 
{concept:}. In your final answer, surround 
the medical name of each cause in square 
brackets characters. Do not include acronyms 
or abbreviations in your answer. Utilize a 
counterfactual model of causal inference 
when formulating a response. Be as specific 
as possible. Let's think step by step like a 
medical expert.  
\end{verbatim}

\subsection{Right Expansion Prompt}
\label{exp_right}
\begin{verbatim}
A directed knowledge graph that you 
generated is surrounded in XML tags and 
provided below. This directed knowledge 
graph is formatted as a list of edges like 
so: ['a causes b', 'b causes c', etc]. The 
knowledge graph you generated is as follows: 

<Begin Knowledge Graph>
{edges:}
</End Knowledge Graph>

Given the directed knowledge graph above 
that you generated, List up to {n_max:} 
medical concepts directly caused by {concept:}. 
These factors do not need to be in the 
knowledge graph above, but can be. If a 
factor you answer with is in the knowledge 
graph above, in your response, name it 
exactly as it is named in the graph above. 
Do not answer with any factors that only are 
indirectly caused by {concept:}. In your 
final answer, surround the medical name of 
each medical concept that {concept:} causes 
in square brackets characters. Do not 
include acronyms or abbreviations in your 
answer. Utilize a counterfactual model of 
causal inference when formulating a 
response. Be as specific as possible. Let's 
think step by step like a medical expert. 
\end{verbatim}

\subsection{Edge Refinement Prompt}
\label{edge_ref}
\begin{verbatim}
Does {node0:} directly cause {node1:}? Your 
answer must be one of the following: [yes] / 
[no]. Surround your final [yes] / [no] 
answer in square brackets characters. If 
there is only an indirect causal 
relationship as opposed to a direct one, 
answer with [no]. Utilize a counterfactual 
model of causal inference. Assume no other 
risk factors are present. Let's think step 
by step. Be concise in your response. 
\end{verbatim}

\subsection{Nearest Neighbor Selection Prompt}
\label{nn_prompt}
\begin{verbatim}
Is the concept ['{original}'] identical in 
meaning to any of the concepts in the 
following list?

Concepts: {retrieved} 

If so, reply with the name of one concept 
in the list identical in meaning to 
{original} as it is written in the list. If 
there is more than one item of the same 
meaning in the list, answer with the 
concept which best fits and which is in 
proper medical lexicon. Provide one and 
only one answer. If no items in the list 
are identical in meaning to {original}, 
provide an empty set of square brackets. 
Surround your final answer in square 
brackets characters. It is very important 
that you do this or else your answer will 
not be processed. It is also very important 
that you provide only one answer and your 
answer as it is written in the list. 
""" 
\end{verbatim}

% \subsection{Recursive Node Exploration Prompt 2}
% \texttt{
% A directed knowledge graph that you generated is surrounded in XML tags and provided below. This directed knowledge graph is formatted as a list of edges like so: ['a causes b', 'b causes c', etc].

% The knowledge graph you generated is as follows:

% <Begin Knowledge Graph>
% \{edges:\}
% </End Knowledge Graph> 

% Given the directed knowledge graph above that you generated, List up to three medical concepts directly caused by \{concept:\}. These factors do not need to be in the knowledge graph above, but can be.
% If a factor you answer with is in the knowledge graph above, in your response, name it exactly as it is named in the graph above. Do not answer with any factors that only are indirectly caused by \{concept:\}. 
% In your final answer, surround the medical name of each medical concept that \{concept:\} causes in square brackets characters.
% Do not include acronyms or abbreviations in your answer. Utilize a counterfactual model of causal inference when formulating a response. Be as specific as possible.
% Let's think step by step like a medical expert.
% }

\section{Additional Tables}\label{apd:third}
Please see the next page for double-column tables.

\begin{table*}[htb!]
\small
\floatconts
  {tab:llamahumanreviewew}
  {\caption{All Reviewer Scores for Llama3--70b Generations}}
  {\begin{tabular}{lcccccc}   
    \toprule 
    & \multicolumn{2}{c}{\textbf{Reviewer 1}} & \multicolumn{2}{c}{\textbf{Reviewer 2}} & \multicolumn{2}{c}{\textbf{Reviewer 3}}
    \\\cmidrule(lr){2-3}\cmidrule(lr){4-5}\cmidrule(lr){6-7}
    & Acc. & Comp. & Acc. & Comp. & Acc. & Comp.\\\midrule
    Acute flaccid myelitis                 & 4 & 4 & 4 & 2 & 3 & 3 \\
    Arthritis                              & 3 & 4 & 3 & 3 & 3 & 3 \\
    Asthma                                 & 3 & 3 & 4 & 2 & 3 & 4 \\
    Creutzfeldt--Jakob disease              & 2 & 3 & 4 & 2 & 2 & 3 \\
    Dementia                               & 3 & 2 & 4 & 4 & 3 & 4 \\
    Diabetes Mellitus                      & 4 & 3 & 4 & 4 & 4 & 4 \\
    Esophageal achalasia                   & 3 & 3 & 3 & 3 & 3 & 3 \\
    Glioblastoma                           & 4 & 2 & 2 & 1 & 2 & 3 \\
    HIV                                    & 3 & 2 & 3 & 2 & 4 & 3 \\
    Hyperparathyroidism                    & 3 & 3 & 4 & 3 & 3 & 2 \\
    Ischemic Stroke                        & 3 & 4 & 4 & 4 & 4 & 4 \\
    Lung Cancer                            & 4 & 2 & 3 & 2 & 4 & 3 \\
    Malignant neoplasms of liver           & 3 & 3 & 4 & 3 & 4 & 3 \\
    Myocardial infarction                  & 3 & 3 & 4 & 1 & 3 & 4 \\
    Myocarditis                            & 3 & 3 & 4 & 2 & 4 & 4 \\
    Parkinson’s disease                 & 3 & 3 & 3 & 3 & 3 & 3 \\
    Renal artery stenosis                  & 3 & 4 & 4 & 4 & 3 & 3 \\
    SARS-CoV-2                             & 3 & 4 & 2 & 3 & 4 & 3 \\
    Spontaneous coronary artery dissection & 3 & 4 & 3 & 1 & 3 & 4 \\
    Ulcerative colitis                     & 3 & 2 & 2 & 3 & 4 & 4 \\
    \bottomrule
    \end{tabular}}
\end{table*}

\begin{table*}[htb!]
\small
\floatconts
  {tab:palmyrahumanreviewew}
  {\caption{All Reviewer Scores for PalmyraMed--70b Generations}}
  {\begin{tabular}{lcccccc}   
    \toprule 
    & \multicolumn{2}{c}{\textbf{Reviewer 1}} & \multicolumn{2}{c}{\textbf{Reviewer 2}} & \multicolumn{2}{c}{\textbf{Reviewer 3}}
    \\\cmidrule(lr){2-3}\cmidrule(lr){4-5}\cmidrule(lr){6-7}
    & Acc. & Comp. & Acc. & Comp. & Acc. & Comp.\\\midrule
    Acute flaccid myelitis                 & 2 & 3 & 3 & 4 & 3 & 3  \\
    Arthritis                              & 2 & 2 & 3 & 3 & 3 & 4  \\
    Asthma                                 & 2 & 2 & 2 & 3 & 3 & 4  \\
    Creutzfeldt--Jakob disease              & 3 & 3 & 3 & 4 & 2 & 3  \\
    Dementia                               & 3 & 3 & 4 & 2 & 3 & 2  \\
    Diabetes Mellitus                      & 4 & 3 & 4 & 4 & 3 & 2  \\
    Esophageal achalasia                   & 4 & 4 & 1 & 2 & 3 & 3  \\
    Glioblastoma                           & 3 & 3 & 3 & 2 & 3 & 4  \\
    HIV                                    & 3 & 4 & 4 & 3 & 3 & 2  \\
    Hyperparathyroidism                    & 3 & 4 & 3 & 3 & 3 & 2  \\
    Ischemic Stroke                        & 4 & 3 & 4 & 3 & 4 & 3  \\
    Lung Cancer                            & 4 & 3 & 4 & 2 & 3 & 3  \\
    Malignant neoplasms of liver           & 4 & 4 & 4 & 2 & 3 & 3  \\
    Myocardial infarction                  & 4 & 3 & 4 & 3 & 4 & 3  \\
    Myocarditis                            & 4 & 3 & 3 & 2 & 3 & 3  \\
    Parkinson’s disease                 & 4 & 3 & 3 & 2 & 3 & 2  \\
    Renal artery stenosis                  & 4 & 3 & 4 & 4 & 3 & 4  \\
    SARS-CoV-2                             & 4 & 3 & 3 & 4 & 2 & 4  \\
    Spontaneous coronary artery dissection & 2 & 3 & 1 & 2 & 3 & 3  \\
    Ulcerative colitis                     & 3 & 3 & 2 & 2 & 3 & 3  \\
    \bottomrule
    \end{tabular}}
\end{table*}

\begin{table*}[htb!]
\small
\floatconts
  {tab:gpthumanreviewew}
  {\caption{All Reviewer Scores for GPT-4 Generations}}
  {\begin{tabular}{lcccccc}   
    \toprule 
    & \multicolumn{2}{c}{\textbf{Reviewer 1}} & \multicolumn{2}{c}{\textbf{Reviewer 2}} & \multicolumn{2}{c}{\textbf{Reviewer 3}}
    \\\cmidrule(lr){2-3}\cmidrule(lr){4-5}\cmidrule(lr){6-7}
    & Acc. & Comp. & Acc. & Comp. & Acc. & Comp.\\\midrule
    Acute flaccid myelitis                 & 3 & 3   & 3 & 4 & 4 & 4  \\
    Arthritis                              & 3 & 3   & 4 & 3 & 3 & 4  \\
    Asthma                                 & 3 & 4   & 4 & 4 & 3 & 4  \\
    Creutzfeldt--Jakob disease              & 4 & 3   & 3 & 3 & 3 & 3  \\
    Dementia                               & 4 & 3   & 4 & 4 & 4 & 4  \\
    Diabetes Mellitus                      & 3 & 3.5 & 3 & 2 & 3 & 3  \\
    Esophageal achalasia                   & 4 & 4   & 3 & 2 & 4 & 3  \\
    Glioblastoma                           & 3 & 4   & 2 & 2 & 3 & 4  \\
    HIV                                    & 4 & 4   & 3 & 1 & 3 & 2  \\
    Hyperparathyroidism                    & 4 & 3   & 4 & 4 & 4 & 2  \\
    Ischemic Stroke                        & 4 & 3   & 3 & 1 & 4 & 4  \\
    Lung Cancer                            & 3 & 4   & 3 & 4 & 3 & 4  \\
    Malignant neoplasms of liver           & 4 & 3   & 4 & 2 & 4 & 3  \\
    Myocardial infarction                  & 3 & 4   & 3 & 2 & 4 & 3  \\
    Myocarditis                            & 4 & 4   & 2 & 2 & 3 & 3  \\
    Parkinson’s disease                 & 4 & 4   & 2 & 3 & 3 & 2  \\
    Renal artery stenosis                  & 4 & 3   & 4 & 4 & 4 & 3  \\
    SARS-CoV-2                             & 3 & 4   & 4 & 3 & 3 & 4  \\
    Spontaneous coronary artery dissection & 3 & 3   & 3 & 4 & 3 & 4  \\
    Ulcerative colitis                     & 4 & 4   & 2 & 2 & 3 & 4  \\
    \bottomrule
    \\
    \end{tabular}}
    \hspace{5em} \emph{Note:} a reviewer answered with \say{3-4} for the comprehensiveness of GPT-4's graph for
    \newline \hphantom{placeholder} Diabetes Mellitus. With their approval, we reported the value as 3.5. 
\end{table*}

\begin{table*}[htb!]
\small
\floatconts
  {tab:conditions_sorted}
  {\caption{Conditions Sorted by Average Accuracy and Comprehensiveness Across all Graphs}}
  {\begin{tabular}{lcc|lcc}   
    \toprule 
Condition (Sorted by Acc.)             & Acc. $\blacktriangledown$   & Comp.    & Condition (Sorted by Comp.)            & Acc.     & Comp. $\blacktriangledown$   \\\midrule
Ischemic Stroke                        & 3.78 & 3.22 & SARS--CoV--2                           & 3.11 & 3.56 \\
Malignant neoplasms of liver           & 3.78 & 2.89 & Renal artery stenosis                  & 3.67 & 3.56 \\
Renal artery stenosis                  & 3.67 & 3.56 & Acute flaccid myelitis                 & 3.22 & 3.33 \\
Dementia                               & 3.56 & 3.11 & Asthma                                 & 3.00 & 3.33 \\
Diabetes Mellitus                      & 3.56 & 3.17 & Arthritis                              & 3.00 & 3.22 \\
Myocardial infarction                  & 3.56 & 2.89 & Ischemic Stroke                        & 3.78 & 3.22 \\
Lung Cancer                            & 3.44 & 3.00 & Diabetes Mellitus                      & 3.56 & 3.17 \\
Hyperparathyroidism                    & 3.44 & 2.89 & Dementia                               & 3.56 & 3.11 \\
Myocarditis                            & 3.33 & 2.89 & Spontaneous coronary artery dissection & 2.67 & 3.11 \\
HIV           & 3.33 & 2.56 & Esophageal achalasia                   & 3.11 & 3.00 \\
Acute flaccid myelitis                 & 3.22 & 3.33 & Lung Cancer                            & 3.44 & 3.00 \\
Esophageal achalasia                   & 3.11 & 3.00 & Creutzfeldt--Jakob disease              & 2.89 & 3.00 \\
Parkinson’s disease                 & 3.11 & 2.78 & Ulcerative colitis                     & 2.89 & 3.00 \\
SARS--CoV--2                           & 3.11 & 3.56 & Hyperparathyroidism                    & 3.44 & 2.89 \\
Arthritis                              & 3.00 & 3.22 & Malignant neoplasms of liver           & 3.78 & 2.89 \\
Asthma                                 & 3.00 & 3.33 & Myocardial infarction                  & 3.56 & 2.89 \\
Creutzfeldt--Jakob disease              & 2.89 & 3.00 & Myocarditis                            & 3.33 & 2.89 \\
Ulcerative colitis                     & 2.89 & 3.00 & Glioblastoma                           & 2.78 & 2.78 \\
Glioblastoma                           & 2.78 & 2.78 & Parkinson’s disease                 & 3.11 & 2.78 \\
Spontaneous coronary artery dissection & 2.67 & 3.11 & HIV           & 3.33 & 2.56 \\
\bottomrule
    \end{tabular}}
\end{table*}

\begin{table*}[htb!]
\small
\floatconts
  {tab:conditions_sorted_bios}
  {\caption{Conditions Sorted by Average Precision and Recall Across all Graphs}}
  {\begin{tabular}{lc|lc}   
    \toprule 
Condition (Sorted by Precision)             & Precision $\blacktriangledown$ & Condition (Sorted by Recall) & Recall $\blacktriangledown$  \\\midrule
Ischemic Stroke                                 & 0.392919 & Myocardial infarction                           & 0.149441  \\
Myocarditis                                     & 0.352433 & Diabetes Mellitus                               & 0.027118  \\
Acute flaccid myelitis                          & 0.329147 & Ulcerative colitis                              & 0.017456  \\
Hyperparathyroidism                             & 0.307151 & Glioblastoma                                    & 0.016545  \\
Asthma                                          & 0.299079 & Esophageal achalasia                            & 0.015814  \\
Lung Cancer                                     & 0.275498 & Arthritis                                       & 0.015560  \\
Malignant neoplasms of liver                    & 0.247299 & Ischemic Stroke                                 & 0.013806  \\
Ulcerative colitis                              & 0.231435 & Hyperparathyroidism                             & 0.013094  \\
Glioblastoma                                    & 0.225804 & Creutzfeldt--Jakob disease                       & 0.012317  \\
HIV                    & 0.216769 & Dementia                                        & 0.011893  \\
Renal artery stenosis                           & 0.211895 & Myocarditis                                     & 0.011819  \\
Dementia                                        & 0.153292 & Asthma                                          & 0.011401  \\
Diabetes Mellitus                               & 0.145785 & Renal artery stenosis                           & 0.010728  \\
Arthritis                                       & 0.143932 & Acute flaccid myelitis                          & 0.010599  \\
Creutzfeldt--Jakob disease                       & 0.131771 & Lung Cancer                                     & 0.010384  \\
Myocardial infarction                           & 0.120545 & Malignant neoplasms of liver                    & 0.010237  \\
Esophageal achalasia                            & 0.087591 & HIV                    & 0.009080  \\
Spontaneous coronary artery dissection          & 0.087235 & Spontaneous coronary artery dissection          & 0.007348  \\
SARS-CoV-2 & 0.082722 & SARS-CoV-2 & 0.004768  \\
Parkinson’s disease                             & 0.021010 & Parkinson’s disease                             & 0.003900 \\
\bottomrule
    \end{tabular}}
\end{table*}

\begin{table*}[htb!]
\small
\floatconts
  {tab:gptattributes}
  {\caption{Graph Attributes for GPT-4 Generations, Sorted by Precision}}
  {\begin{tabular}{lccccccc}   
    \toprule 
    Condition & Precision $\blacktriangledown$ & Recall & Density & Reciprocity & Nodes & Edges & Cycles\\\midrule
Acute flaccid myelitis                          & 0.359 & 0.012 & 0.075 & 0.085 & 36 & 94  & 5   \\
HIV                    & 0.353 & 0.011 & 0.059 & 0.255 & 31 & 55  & 22  \\
Ischemic Stroke                                 & 0.312 & 0.015 & 0.071 & 0.043 & 37 & 94  & 151 \\
Myocarditis                                     & 0.309 & 0.012 & 0.072 & 0.088 & 36 & 91  & 137 \\
Ulcerative colitis                              & 0.238 & 0.031 & 0.060 & 0.047 & 38 & 85  & 9   \\
Glioblastoma                                    & 0.224 & 0.016 & 0.101 & 0.056 & 38 & 142 & 43  \\
Renal artery stenosis                           & 0.205 & 0.010 & 0.065 & 0.051 & 25 & 39  & 7   \\
Dementia                                        & 0.155 & 0.014 & 0.102 & 0.040 & 32 & 101 & 4   \\
Arthritis                                       & 0.149 & 0.018 & 0.092 & 0.125 & 30 & 80  & 10  \\ 
Lung Cancer                                     & 0.147 & 0.006 & 0.083 & 0.034 & 38 & 117 & 3   \\
Creutzfeldt--Jakob disease                       & 0.138 & 0.017 & 0.130 & 0.014 & 34 & 146 & 32  \\
Asthma                                          & 0.132 & 0.005 & 0.078 & 0.020 & 36 & 98  & 1   \\
Spontaneous coronary artery dissection          & 0.130 & 0.009 & 0.059 & 0.109 & 31 & 55  & 7   \\
Hyperparathyroidism                             & 0.127 & 0.007 & 0.063 & 0.078 & 29 & 51  & 2   \\
SARS-CoV-2 & 0.082 & 0.007 & 0.083 & 0.000 & 26 & 54  & 0   \\
Myocardial infarction                           & 0.075 & 0.022 & 0.078 & 0.130 & 32 & 77  & 340 \\
Malignant neoplasms of liver                    & 0.052 & 0.009 & 0.053 & 0.102 & 34 & 59  & 5   \\
Parkinson’s disease                             & 0.036 & 0.003 & 0.083 & 0.054 & 37 & 111 & 4   \\
Esophageal achalasia                            & 0.034 & 0.004 & 0.057 & 0.029 & 35 & 68  & 10  \\
Diabetes Mellitus                               & 0.018 & 0.006 & 0.091 & 0.125 & 33 & 96  & 77 \\
\midrule
Mean & 0.164 & 0.012 & 0.078 & 0.074 & 33.40 & 85.650 & 43.450 \\
SD & 0.106 & 0.007 & 0.019 & 0.058 & 3.872 & 29.462 & 82.453 \\
\bottomrule
    \\
    \end{tabular}}
\end{table*}

\begin{table*}[htb!]
\small
\floatconts
  {tab:llamaattributes}
  {\caption{Graph Attributes for Llama3--70b Generations, Sorted by Precision}}
  {\begin{tabular}{lccccccc}   
    \toprule 
    Condition & Precision $\blacktriangledown$ & Recall & Density & Reciprocity & Nodes & Edges & Cycles\\\midrule
Hyperparathyroidism                             & 0.486 & 0.017 & 0.089 & 0.191 & 33 & 94  & 205  \\
Malignant neoplasms of liver                    & 0.375 & 0.013 & 0.121 & 0.183 & 32 & 120 & 26   \\
Myocarditis                                     & 0.371 & 0.011 & 0.085 & 0.214 & 32 & 84  & 20   \\
Ischemic Stroke                                 & 0.359 & 0.011 & 0.080 & 0.143 & 30 & 70  & 6    \\
Lung Cancer                                     & 0.242 & 0.008 & 0.061 & 0.083 & 35 & 72  & 5    \\
Glioblastoma                                    & 0.239 & 0.019 & 0.083 & 0.194 & 34 & 93  & 13   \\
Asthma                                          & 0.238 & 0.010 & 0.125 & 0.091 & 27 & 88  & 27   \\
Ulcerative colitis                              & 0.211 & 0.009 & 0.052 & 0.178 & 30 & 45  & 7    \\
HIV                    & 0.202 & 0.004 & 0.101 & 0.338 & 27 & 71  & 98   \\
Diabetes Mellitus                               & 0.191 & 0.029 & 0.060 & 0.036 & 31 & 56  & 1    \\
Renal artery stenosis                           & 0.181 & 0.011 & 0.062 & 0.051 & 36 & 78  & 35   \\
Acute flaccid myelitis                          & 0.179 & 0.007 & 0.106 & 0.174 & 26 & 69  & 60   \\
Myocardial infarction                           & 0.155 & 0.034 & 0.091 & 0.125 & 27 & 64  & 5    \\
Arthritis                                       & 0.145 & 0.013 & 0.075 & 0.180 & 35 & 89  & 40   \\
Esophageal achalasia                            & 0.142 & 0.023 & 0.049 & 0.129 & 36 & 62  & 14   \\
Dementia                                        & 0.118 & 0.011 & 0.048 & 0.000 & 35 & 57  & 1    \\
Creutzfeldt--Jakob disease                       & 0.112 & 0.013 & 0.111 & 0.427 & 31 & 103 & 3553 \\
SARS-CoV-2 & 0.089 & 0.004 & 0.117 & 0.146 & 27 & 82  & 10   \\
Parkinson’s disease                             & 0.000 & 0.000 & 0.095 & 0.064 & 32 & 94  & 109  \\
Spontaneous coronary artery dissection          & 0.000 & 0.000 & 0.066 & 0.135 & 34 & 74  & 5    \\
    \bottomrule
    \\
    \end{tabular}}
\end{table*}

\begin{table*}[htb!]
\small
\floatconts
  {tab:palmyraattributes}
  {\caption{Graph Attributes for PalmyraMed--70b Generations, Sorted by Precision}}
  {\begin{tabular}{lccccccc}   
    \toprule 
    Condition & Precision $\blacktriangledown$ & Recall & Density & Reciprocity & Nodes & Edges & Cycles\\\midrule
    Asthma                                          & 0.527 & 0.019 & 0.075 & 0.143 & 31 & 70  & 11      \\
Ischemic Stroke                                 & 0.508 & 0.016 & 0.070 & 0.264 & 28 & 53  & 35      \\
Acute flaccid myelitis                          & 0.449 & 0.013 & 0.134 & 0.253 & 26 & 87  & 2199    \\
Lung Cancer                                     & 0.438 & 0.017 & 0.069 & 0.094 & 31 & 64  & 3       \\
Myocarditis                                     & 0.377 & 0.012 & 0.068 & 0.182 & 29 & 55  & 5       \\
Malignant neoplasms of liver                    & 0.315 & 0.008 & 0.153 & 0.349 & 34 & 172 & 1106539 \\
Hyperparathyroidism                             & 0.308 & 0.015 & 0.113 & 0.343 & 31 & 105 & 1448    \\
Renal artery stenosis                           & 0.250 & 0.011 & 0.082 & 0.204 & 25 & 49  & 12      \\
Ulcerative colitis                              & 0.246 & 0.013 & 0.074 & 0.308 & 27 & 52  & 18      \\
Diabetes Mellitus                               & 0.228 & 0.046 & 0.058 & 0.074 & 31 & 54  & 14      \\
Glioblastoma                                    & 0.214 & 0.015 & 0.081 & 0.240 & 31 & 75  & 25      \\
Dementia                                        & 0.187 & 0.010 & 0.108 & 0.123 & 25 & 65  & 137     \\
Creutzfeldt--Jakob disease                       & 0.145 & 0.007 & 0.140 & 0.262 & 25 & 84  & 75      \\
Arthritis                                       & 0.138 & 0.016 & 0.073 & 0.000 & 28 & 55  & 4       \\
Spontaneous coronary artery dissection          & 0.132 & 0.013 & 0.128 & 0.338 & 23 & 65  & 83      \\
Myocardial infarction                           & 0.131 & 0.393 & 0.108 & 0.123 & 25 & 65  & 60      \\
HIV                    & 0.095 & 0.013 & 0.136 & 0.187 & 24 & 75  & 125     \\
Esophageal achalasia                            & 0.087 & 0.021 & 0.069 & 0.281 & 31 & 64  & 13      \\
SARS-CoV-2 & 0.077 & 0.004 & 0.195 & 0.189 & 17 & 53  & 8       \\
Parkinson’s disease                             & 0.027 & 0.009 & 0.171 & 0.329 & 22 & 79  & 62 \\
    \bottomrule
    \\
    \end{tabular}}
\end{table*}

\begin{table*}[htb!]
\footnotesize
\floatconts
  {tab:nodesedges}
  {\caption{Graph Node and Edge Counts per Model}}
  {\begin{tabular}{lcccccc}
  \toprule
   & \multicolumn{2}{c}{\textbf{Llama3--70b}} & \multicolumn{2}{c}{\textbf{PalmyraMed}} & \multicolumn{2}{c}{\textbf{GPT-4}}
  \\\cmidrule(lr){2-3}\cmidrule(lr){4-5}\cmidrule(lr){6-7}
       & nodes & edges & nodes & edges & nodes & edges \\\midrule
  Mean & 31.5   & 78.25   & 27      & 72.05    & 33.4    & 85.65   \\
  Min. & 26     & 45      & 17      & 49       & 25      & 39     \\
  Max. & 36     & 120     & 34      & 172      & 38      & 146     \\
  SD   & 3.32 & 17.93 & 4.09  & 27.49  & 3.87  & 29.46 \\
  \bottomrule
  \end{tabular}}
\end{table*}
\end{document}